\documentclass[sn-apa]{sn-jnl}

\usepackage{graphicx}%
\usepackage{multirow}%
\usepackage{amsmath,amssymb,amsfonts}%
\usepackage{amsthm}%
\usepackage{mathrsfs}%
\usepackage[title]{appendix}%
\usepackage{xcolor}%
\usepackage{textcomp}%
\usepackage{manyfoot}%
\usepackage{booktabs}%
\usepackage{algorithm}%
\usepackage{algorithmicx}%
\usepackage{algpseudocode}%
\usepackage{listings}%
\usepackage{lipsum} 
\usepackage{comment}
\usepackage{footmisc}

\usepackage{tikz}
\usepackage{tikz-3dplot}
\usepackage{tkz-euclide}
\usetikzlibrary{arrows,automata,external,patterns}
\usepackage{pgfplots}
\usepackage{pgfplotstable}
\pgfplotsset{compat=newest}
\usepgfplotslibrary{external} 

\pgfmathdeclarefunction{gauss}{2}{%
    \pgfmathparse{1/(#2*sqrt(2*pi))*exp(-((x-#1)^2)/(2*#2^2))}%
}

\tikzset{
    declare function={
        normcdf(\x,\m,\s)=1 / (1 + exp(-0.07056*((\x-\m)/\s)^3 - 1.5976*(\x-\m)/\s));
    }
}

\excludecomment{hide}

\theoremstyle{thmstyleone}%
\theoremstyle{thmstyletwo}%

\theoremstyle{thmstylethree}%

\begin{document}

\title[Normality and the Turing Test]{Normality and the Turing Test}

\author{\centering Alexandre Kabbach\footnote{
    \href{mailto:alexandre@kabbach.net}{alexandre@kabbach.net}
} \\ Chubu Gakuin University}

\abstract{
    This paper proposes to revisit the Turing test through the concept of \emph{normality}.
    Its core argument is that the Turing test is a test of \emph{normal intelligence} as assessed by a \emph{normal judge}.
    First, in the sense that the Turing test targets normal/average rather than exceptional human intelligence, so that successfully passing the test requires machines to ``make mistakes'' and display imperfect behavior just like normal/average humans.
    Second, in the sense that the Turing test is a statistical test where judgments of intelligence are never carried out by a single ``average'' judge~(understood as non-expert) but always by a full jury.
    As such, the notion of ``average human interrogator'' that Turing talks about in his original paper should be understood primarily as referring to a mathematical abstraction made of the normalized aggregate of individual judgments of multiple judges.
    Its conclusions are twofold.
    First, it argues that large language models such as ChatGPT are unlikely to pass the Turing test as those models precisely target exceptional rather than normal/average human intelligence.
    As such, they constitute models of what it proposes to call \emph{artificial smartness} rather than artificial intelligence, insofar as they deviate from the original goal of Turing for the modeling of artificial minds.
    %
    Second, it argues that the objectivization of normal human behavior in the Turing test fails due to the game configuration of the test which ends up objectivizing \emph{normative ideals} of normal behavior rather than normal behavior \emph{per se}.  
}


\maketitle

\section{Introduction}
\label{sec:intro}

When~\citet{turing1950} introduces his seminal paper in \emph{Mind} on \emph{Computing machinery and intelligence}, he sets the goal of the field of artificial intelligence straight: the aim is to build a machine capable of passing his ``Imitation Game''. 
The Imitation Game\textemdash{}now better know as the ``Turing test''\textemdash{}is an indistinguishibility test.
It consists in having a human interrogator sit alone in a room and communicate  with two separate interlocutors through a text-only channel while having to decide, based solely on the content of their textual interactions, which one of the two participants is the human and which one is the machine.
If, on a repeated number of occasions, the human interrogator cannot distinguish the machine from the other human participants, then the machine can be said to have successfully passed the test. 

Turing intended his test to substitute the question ``Are there imaginable digital computers which would do well in the imitation game?'' for the question ``Can machines think?''\textemdash{}a latter question which he himself considered ``too meaningless to deserve discussion''~\citep[p.442]{turing1950}.
The Turing test thus crucially rests on the assumption that intelligence should be approached as a human \emph{faculty}, a fundamental ability to ``think'' that one possesses by virtue of being human.
Which is why some have argued in the past that it constitutes a test of ``humanity'' rather than intelligence \emph{per se}~\citep{fostel1993}.
\begin{hide}
    In this paper, however, I will argue that the Turing test constitutes a test of \emph{normality} rather than \emph{humanity} strictly speaking.
    My point indeed is that if it treats ``being intelligent'' as being more or less synonymous with ``being human'', the Turing test actually equates intelligent \emph{behavior} with \emph{normal} human behavior rather than human behavior strictly speaking.
\end{hide}

However, if the Turing test treats ``being intelligent'' as being more or less synonymous with ``being human'', does that mean that it equates intelligent \emph{behavior} with human behavior, in turn?  
Not exactly, and this is where the fundamental contribution of this paper lies. 
In this paper indeed, I argue that the Turing test equates intelligent behavior with \emph{normal} human behavior rather than human behavior strictly speaking. 
In short, it is a test of \emph{normality} rather than humanity \emph{per se}.

There are at least two ways to make sense of this concept of ``normality'' for our present purpose.
The first one is through the normal/exceptional dichotomy and by appealing to a normative interpretation of the normal that evokes the average as \emph{mediocre}. 
%
Crucial to the understanding of the Turing test is that it targets \emph{normal/average} rather than \emph{exceptional} human intelligence\textemdash{}the average intelligence of most people rather than the exceptional intelligence of a happy few geniuses~(\S\ref{sec:arg:normint}).
%
%
The normal/average brain, however, is necessarily imperfect and bound to make mistakes at some point, which is why the Turing test requires machines to make mistakes so as to be truly indistinguishable from normal people. 
This explains why, in turn, large language models such as ChatGPT are unlikely to pass the Turing test~(\S\ref{sec:arg:chatgpt}). 
The argument is plain: who needs machines that make mistakes?
For the practical purpose those language models are usually being put to use, normal intelligence is not enough: they need to be \emph{smart}.
They need to conform to a normative ideal of correct and exceptional human behavior that no human being actually abides to in practice.  
In practice indeed, real people make mistakes\textemdash{}they always deviate from whichever normative ideal of correct human behavior they live by\textemdash{}and that is precisely what makes them human.

The second way to make sense of the concept of ``normality'' then is through the statistical interpretation of the normal this time, which evokes the average understood both as the \emph{mean} and the~(statistically) \emph{typical}.
A second and no less crucial aspect of the Turing test to bear in mind is that it is a \emph{statistical} test where judgments of intelligent behavior are never produced by a single human interrogator but always by a full jury~(\S\ref{sec:arg:stat}).
\begin{hide}
    As such, the concept of ``average human interrogator'' mentioned in the original specifications of the Turing test should be understood not so much as referring to a non-expert individual than to a~(mathematical) abstraction corresponding to the standard objectivization practice of the behavioral sciences consisting in averaging individual responses so as to eliminate intrinsic subject variability.
\end{hide}
The notion of ``average interrogator'' that \citet[p.442]{turing1950} talks about in his original paper should thus be understood primarily as an \emph{idealization} referring to a mathematical abstraction consisting in aggregating judgments of individual human interrogators and computing their mean value\textemdash{}a mean value considered to be ``truer'' and more objective than those individual judgments alone.
The methodology of the Turing test is thereby consistent with the standard objectivization practice of the behavioral sciences and with its foundational epistemological assumption: that human variations follow a normal distribution and that individual variability should be treated as \emph{error} around a mean value characterizing a statistical type.

In sum, this paper argues that the Turing test has often been~(and keeps being) critically misunderstood. 
Turing never intended to target what we commonly refer to as ``intelligence'' in everyday language and which actually refers to \emph{exceptional} intelligence most of the time.
What he wanted to investigate was \emph{normal} intelligence understood as \emph{average/typical} human intelligence\textemdash{}in accordance with the then and now dominant normalist paradigm taking normality as its ``central organizing concept''~\citep{hacking1990,hacking1996b}.
By shifting the target of inquiry from the exceptional to the normal, Turing brings back intelligence into normalism in full.  
His Imitation Game takes place within a naturalistic anthropology of human being dedicated to the characterization of the human \emph{type}\textemdash{}or at least \emph{a} human type since the type objectivized by the Turing test will necessarily be both socio-culturally and  socio-historically situated~\citep[see][]{french1990}.
Ultimately, intelligent behavior for Turing is indeed to be understood as referring to \emph{an instance of} normal/typical human behavior.
As we will see, however, this anthropological enterprise collapses against the ``game'' aspect of the test, as it leads to the objectivization not of normal behavior but of a \emph{normative ideal} of normal behavior\textemdash{}not how normal people \emph{do} behave but how they \emph{ought} to behave to be considered ``normal''~(\S\ref{sec:arg:game}).

\begin{hide}
    The rest of the paper proceeds as follows. 
    Section~\ref{sec:def} begins by introducing the core concept of normality alongside Turing's conception of intelligence.
    Section~\ref{sec:def:what} introduces the concept of normality from a historical perspective, distinguishing in particular its \emph{biomedical} interpretation understood as the functional or the healthy from its \emph{statistical} interpretation understood as the average or~(statistically) typical. 
    After detailing the epistemological contribution of the concept of statistical normality and notably the crucial role that it has played in the foundation of the behavioral sciences and of modern psychology, the section shifts to the introduction of two distinct normative interpretations of statistical normality: the normal/average as \emph{good} on one side and the normal/average as only \emph{mediocre} on the other.
    The latter normative interpretation paves the way for the normal/exceptional dichotomy and, through it, to the core argument of this section: that the Turing test actually targets normal/average/mediocre rather than exceptional human intelligence.

    Section~\ref{sec:def:intelligence} introduces Turing's conception of intelligence in more details, tracing back its origins to the notions of \emph{intellect}, \emph{understanding} and the cartesian concept of \emph{reason}.
    It proposes to revisit the distinction between \emph{intelligence} and \emph{intelligent behavior} through the fundamental contribution of \emph{development}, emphasizing that Turing's conception of intelligence as a faculty implies that intelligence is always but a potential that can only become behaviorally manifest through gradual exposure to environmental stimuli.
    After comparing Turing's approach to the faculty of intelligence to Chomsky's approach to the faculty of language, it concludes that the Turing test remains fundamentally normalist in that it does not aim to characterize the boundaries of human cognition\textemdash{}determining all that can and cannot be human\textemdash{}but only \emph{one} of its possible manifestation: the intelligent behavior associated with a particular community of human beings. 
    Such considerations explain, in turn, why the Turing test cannot be taken to provide an operational characterization of intelligence\textemdash{}i.e., providing both a \emph{necessary} and \emph{sufficient} condition for the possession of human intelligence. 
    Considering otherwise would be like saying that manifesting a particular type of developmentally-situated linguistic behavior\textemdash{}say, speaking English\textemdash{}provides both a necessary and sufficient condition for the possession of the faculty of language, which everyone will agree is absurd~(at least for the necessary part).

    Section~\ref{sec:smart} revisits the Turing test through the normal/exceptional dichotomy. 
    Section~\ref{sec:smart:error} introduces the fundamental distinction between \emph{intelligence} and \emph{smartness} focusing on the opposition between intelligent behavior understood as normal/average human behavior on one side and smart behavior understood as exceptional human behavior on the other.
    The notion of smart behavior is further refined as a \emph{normative ideal} of \emph{correct} in addition to \emph{exceptional} human behavior, which enables two things. 
    First, framing smart behavior as a \emph{normative ideal} of human behavior explains how smartness characterizes an \emph{ideal} type of human behavior from which most if not all real/normal people deviate. 
    Second, framing smart behavior as a normative ideal of \emph{correct} human behavior explains why the Turing test, as a test of normal/average human behavior, specifically requires machines to \emph{make mistakes} and deviate from the normative ideal of smart behavior in the way normal people would.
    The intelligence/smartness dichotomy is then illustrated through the example of \emph{spelling} as perfect spelling is taken\textemdash{}in English at least\textemdash{}to constitute a prototypical example of a normative ideal on human behavior that normal people ought to abide to but never quite manage to in practice.
    The example provides a concrete way to illustrate the fundamental difference between the two scientific projects that constitute \emph{artificial smartness} on one side and \emph{artificial intelligence} on the other: the former being dedicated to the modeling of correct and exceptional human behavior in general and perfect/correct spelling in particular while the latter is dedicated\textemdash{}for Turing at least\textemdash{}to the modeling of normal/average human behavior in general and imperfect/normal/average spelling in particular.

    With those distinctions in mind, Section~\ref{sec:smart:chatgpt} introduces the core implications of the normalist reading of the Turing test for large language models such as ChatGPT.
    Its main argument is that such models are unlikely to pass the Turing test as they are models of artificial smartness rather than artificial intelligence\textemdash{}targeting exceptional rather than normal human intelligence. 
    The argument however, is analytical and not ontological. 
    That is, its point is not to claim that such models could \emph{never} be made so as to successfully pass the Turing test but only that, for the practical applications those models are usually being put to use, it makes more sense for them to model smart/exceptional/correct human behavior rather than intelligent/normal/incorrect human behavior. 
    The argument is plain: \emph{who needs machines that make mistakes?}
    That being said, if nothing prevents us from training those very same models on new data so as to learn how to ``make mistakes'', what is crucial to understand is that \emph{not any mistakes goes} as far as the Turing test is concerned: to successfully pass the Turing test, those models would need to produce \emph{normal/average mistakes} as normal people would. 
    And determining the nature of those normal mistakes\textemdash{}and creating the corresponding datasets along with it\textemdash{}is quite a scientific challenge. 
    It is, in fact, the very challenge that \emph{artificial intelligence} as originally conceived by Turing was supposed to tackle.

    Section~\ref{sec:type} moves on to the understanding of normality as a statistical type.
    Section~\ref{sec:type:stat} reminds us that the Turing test is a statistical test where success is never assessed by a single judge by always by a full jury. 
    %
    %
    %
    Understanding that the Turing test characterizes a test of \emph{normality} rather than \emph{humanity} strictly speaking then enables two things.
    First, it explains how judgments of ``humanity'' can vary significantly from one judge to the next: human interrogators are simply expressing what feels normal \emph{to them}.
    Second, it provides a deeper understanding of the nature of the objectivization process in the Turing test, made possible by the use of statistics: the point is to move past the normal as \emph{familiar} to reach the \emph{statistically} normal\textemdash{}moving away from a form of subjective normality so as to characterize a purportedly objective statistical \emph{type}. 
    Such considerations then beg the question of \emph{which type} will be objectivized by the Turing test, to which the joint analyses of the specifications of the test, the methodological biases of experimental psychology and the historical roots of statistical normality suggests that it is unlikely to be a universal or even a representative human type. 
    \begin{hide}
        That being said, it is true that Turing tells us nothing about how the populations of human participants and interrogators should be sampled, which begs the question of which type will be objectivized. 
    \end{hide}

    Section~\ref{sec:type:limit} finishes on some important limitations of the Turing test, starting with the ``game'' aspect of the test which puts participants in a configuration where they compete to appear \emph{as normal as possible} from the perspective of the average human interrogator. 
    The problem with that aspect of the test is that it leads human participants to behave not as they would normally do but as they think they \emph{ought} to so as to appear normal in that context\textemdash{}ultimately leading to the objectivization not of normal human behavior \emph{per se} but of a \emph{normative ideal} of normal human behavior.
    The second limitation of the test pertains to the implicit prerequisite of ``cultural alignment'' between human interrogators and participants\textemdash{}requiring them both to share roughly the same conception of what it means to be ``normal''\textemdash{}and which conditions the meaningfulness of the test results themselves.
    The problem with that prerequisite is that it is impossible to enforce in practice given the fundamentally open-ended nature of the test, so that it is ultimately impossible to guarantee \emph{both} a completely open-ended \emph{and} meaningful Turing test. 
    The next limitations pertain less to the Turing test itself than to its overarching normalist paradigm.
    The third limitation points at the descriptive/prescriptive tension behind the concept of normality\textemdash{}constantly navigating between what \emph{is} and what \emph{ought to be}\textemdash{}and questions whether what is objectivized by the Turing test can be considered ``real'' in any meaningful sense of the term or whether it is not simply a constructed fiction. 
    This descriptive/prescriptive tension, in turn, is what makes normality a \emph{constantly moving target} which questions, in the case of artificial intelligence, whether we are not actually modifying our target of inquiry~(intelligence) as we attempt to model it. 
    The final limitation of the test is that of the concept of statistical normality itself, which proves of little help when the goal is to determine the nature of the human faculty of intelligence.  
    The point is that human mind is simply not reducible to the normal/average/typical mind, as there is simply more to being human than to being normal.
\end{hide}

\section{Normality and intelligence}
\label{sec:def}

\subsection{What is normal?}
\label{sec:def:what}
\begin{hide}
    x. check quetelet notes
    x. check galton notes 
    x. check methodology notes 
    x. check statistics notes
    x. check malabou notes
    x. check intelligence notes 
    x. check typical notes 
    x. check hacking notes
    x. check metrological-realism notes
\end{hide}

\begin{hide}
    Section~\ref{sec:def} begins by introducing the core concept of normality alongside Turing's conception of intelligence.
    Section~\ref{sec:def:what} introduces the concept of normality from a historical perspective, distinguishing in particular its \emph{biomedical} interpretation understood as the functional or the healthy from its \emph{statistical} interpretation understood as the average or~(statistically) typical. 
    After detailing the epistemological contribution of the concept of statistical normality and notably the crucial role that it has played in the foundation of the behavioral sciences and of modern psychology, the section shifts to the introduction of two distinct normative interpretations of statistical normality: the normal/average as \emph{good} on one side and the normal/average as only \emph{mediocre} on the other.
    The latter normative interpretation paves the way for the normal/exceptional dichotomy and, through it, to the core argument of this section: that the Turing test actually targets normal/average/mediocre rather than exceptional human intelligence.
\end{hide}

\begin{quote}
    Normality is $[\ldots]$
    both timeless and dated, an idea that in some sense has been with us always, but which can in a moment adopt a completely new form of life. 
    $[\ldots]$
    As a word, `normal' $[\ldots]$
    acquired its present most common meaning only in the 1820s. 
    $[\ldots]$
    The normal was one of a pair. 
    Its opposite was the pathological and for a short time its domain was chiefly medical. 
    Then it moved into the sphere of\textemdash{}almost everything. 
    People, behaviour, states of affairs, diplomatic relations, molecules: all these may be normal or abnormal. 
    The word became indispensable because it created a way to be `objective' about human beings.
    The word is also like a faithful retainer, a voice from the past. It uses a power as old as Aristotle to bridge the fact/value distinction, whispering in your ear that what is normal is also all right.~\citep[p.160]{hacking1990} 
\end{quote}
There are at least two ways to understand the concept of normality.
The first one, as the quote from Hacking above suggests, is through its \emph{biomedical} interpretation where the normal evokes the functional or the healthy by opposition to the dysfunctional or the unhealthy\textemdash{}as in the famous ``normal/pathological'' dichotomy of \citet{canguilhem1966}. 
In this context, normality is typically associated with the concept of~(biological) \emph{function}: the normal, \citet[p.494]{king1945} tells us, is ``that which functions in accordance with its design''.
The \emph{normal heart}, for instance, is that which can adequately satisfy its function to \emph{pump blood}.

As far as intelligence is concerned, however, it is not so much this biomedical interpretation than the \emph{statistical} interpretation of the normal that will interest us, where it evokes the standard, the typical, the frequent, the usual, the common and of course, the \emph{average}.
Those two interpretations may occasionally overlap, of course, but they are also crucially distinct: ``The average may be, and very often is, abnormal'' \citet[p.493]{king1945} emphasizes and indeed, an average heart may not always correspond to a functional heart and \emph{vice-versa}~\citep[see][p.580]{wachbroit1994}.

Historically, the concept of statistical normality has been associated with two distinct normative interpretations~\citep[see][ch.19\textendash21, for details]{hacking1990}. 
\begin{figure}[htbp]
    \begin{center}
        \begin{minipage}{0.44\textwidth}
            \centering
            \resizebox{\linewidth}{!}{
                \begin{tikzpicture}
                    \begin{axis}[
                        ticks=none,
                        samples=100,
                        yticklabel=\empty,
                        xticklabel=\empty
                        ]
                        \addplot [thick] {gauss(0,1.5)};
                        \draw [dashed] (0,0 |- current axis.south) -- (0,0 |- current axis.north);
                    \end{axis}
                \end{tikzpicture}
            }
            \caption{The \emph{standard} normal curve, centered on the mean/average~(dashed line).}
            \label{fig:normal:standard}
        \end{minipage}
        \hspace{.1\textwidth}
        \begin{minipage}{0.44\textwidth}
            \centering
            \resizebox{\linewidth}{!}{
                \begin{tikzpicture}
                    \begin{axis}[
                        ticks=none,
                        samples=100,
                        yticklabel=\empty,
                        xticklabel=\empty
                        ]
                        \addplot [thick] {normcdf(x,0,1.5)};
                        \draw [dashed] (0,0 |- current axis.south) -- (0,0 |- current axis.north);
                    \end{axis}
                \end{tikzpicture}
            }
            \caption{The \emph{cumulative} normal curve, centered on the mean/average~(dashed line).}
            \label{fig:normal:cumulative}
        \end{minipage}
    \end{center}
\end{figure}
The first one\textemdash{}where the average evokes the correct, the good, the right or the ideal\textemdash{}is the one typically associated with the standard representation of the normal curve in Figure~\ref{fig:normal:standard},  historically known as the astronomers' ``error law''~(or ``law of errors''), the ``Gaussian distribution'', the ``bell curve'' and of course, the ``\emph{normal} distribution'', the ``\emph{normal} law'' or the ``\emph{normal} curve''.
It is the normative interpretation most often associated with the Belgian astronomer turned statistician Adolphe Quetelet and with his theory of the ``Average Man'' in particular\textemdash{}corresponding to an ideal human being made of the aggregate of all average human attributes or characteristics, both physical and moral~\citep{quetelet1835e}.\footnote{
    See also~\citep{donnelly2015} for details.
}
Under this normative interpretation, \emph{every} deviation from the normal/average is considered undesirable\textemdash{}as with the typical example of the ``Body Mass Index''~(BMI) originally called the ``Quetelet Index''~\citep[see][]{eknoyan2008}.
As far as the BMI is concerned, you are  either ``underweight'' if deviating leftwards from the average, or ``overweight'' if deviating rightwards. 
Only in the average dashed line Figure~\ref{fig:normal:standard} is your weight considered perfectly normal, and so perfectly \emph{ideal}.

\begin{hide}
    Quetelet had a tremendous influence on modern psychology and on the behavioral sciences at large~\citep[see][for overviews]{jahoda2015,tafreshi2022}.
    Many of his contributions have now made their way into modern Science as simply ``good practice''\textemdash{}such as the removal of statistical outliers or the averaging of individual responses to experimental conditions~\citep[see][especially ch.5 on \emph{The triumph of the aggregate}, for an overview]{danziger1990}.
    Such methodological practices, however, remain grounded in a fundamental epistemological assumption, namely, that the normal/average ought to become to proper target of scientific inquiry, for only it characterizes something \emph{objective} as far as human being is concerned.
    This epistemological assumption is motivated by a fundamental empirical observation, in turn: that many human attributes~(e.g. height, weight) follow a normal distribution across populations.
    Those considerations constitute the theoretical underpinnings of Quetelet's theory of the ``Average Man''.

    Quetelet's scientific practice was crucially inspired by the metrological realism of the natural sciences of his time and of astronomy in particular\textemdash{}remember that he was an astronomer himself\textemdash{}which lead him to argue that individual variations could be treated as ``error'' around a ``true'' mean or average~\citep{porter1985}.
    \begin{hide}
        In his paper on \emph{The Meaning of Normal}, \citet[pp.498\textendash499]{king1945} explains how the word ``average'' derives from two Latin words\textemdash{}the preposition \emph{ad} and the noun \emph{verum}\textemdash{}the combination of which means ``toward the truth'' or ``approaching the truth''.
    \end{hide}
    Astronomers had indeed long considered that individual measures of celestial objects necessarily contained some amount of error, but that such error could be mitigated by retaining only the \emph{average} measure of their aggregates\textemdash{}an average measure supposed to capture the \emph{true} measure of an otherwise \emph{real} entity~\citep[see also][for a comprehensive overview]{stigler1986}.
    Quetelet's fundamental contribution to the Science of his time was to borrow astronomy's method of averages but to apply it to \emph{people} this time~\citep[ch.1]{rose2016}. 
    This gave us the now dominant paradigm for the study of human being, a paradigm dedicated to the study of normal responses, normal behaviors, normal brains and normal people in general and which I propose to call the \emph{paradigm of normalism}.\footnote{
        Though historically associated with psychology and the behavioral sciences, normalism has since spread into other fields such as neuroscience. 
        \citet{rose2016} details indeed how the field of neuroscience remains fundamentally grounded in Quetelet's epistemological assumptions, insofar as it is dedicated to the study of an ``Average Brain''\textemdash{}paralleling the ``Average Man'' of Quetelet\textemdash{}quite literally made of an aggregate of individual brains, in the sense that every reported brain activity is always constituted of an averaging of individual brain imaging. 
        %
        That Average Brain, in turn, is presumed to represent ``the normal, typical brain, while each individual brain represents a variant of this normal brain''~\citep[p.20]{rose2016}.
    }
    Normalism marked a radical turning point in the study of human being in the nineteenth century, as it ``displaced the Enlightenment idea of human nature as a central organizing concept'' to focus on ``a model of normal people with laws of dispersion''~\citep[][pp.vii;xi]{hacking1990}.\footnote{
        See also~\citep{hacking1996b}, especially for its explicit mention of the study of intelligence under normalism.
    } 
    Under this normalist paradigm, the concept of normality became a ``metaconcept'' that ``provides a way of thinking that is applied to human beings\textemdash{}their actions and their societies\textemdash{}in every possible aspect'' through ``a completely unifying theme and even methodology''~\citep[p.64]{hacking1996b}.
    \begin{hide}
        \citep{hacking1996b}
        \begin{quote}
            \textbf{What we might call Human-Nature-Thinking is assuredly one highbrow mode of thought, shared by all the memorable figures of the Enlightenment.} It was at the core of their moral philosophy. It underlay their vision of rationality. Now one of the themes of the editors' organization of this volume is a contrast between modem and postmodern, with rationality, abstraction, theorizing, and unification of knowledge being a feature of the modem. I do not myself much use those words,``modem'' and ``postmodern,'' so popular at present and clearly so helpful to many other thinkers. But here goes. I do think that the modem is all too often described by supposedly postmodern thinkers in a disastrously``modem'' and totalizing way, as if our postmodems were so oedipally transfixed that they regress to modernism when thinking about modernism. The earlier era of human nature and the utterly different present era of normal people are both presented, by our thoroughly modem postmodernists, as part of what they call the modem. \textbf{It is true that the organizing concepts of human nature and of normal people are both hegemonic. Each provides a way of thinking that is applied to human beings their actions and their societies in every possible aspect. Each provides a completely unifying theme and even methodology. Each in short is what is currently called modern. They are nevertheless completely different ways of thinking, and we smudge and blend them at our peril.}
        \end{quote}
    \end{hide}
    A central argument of this paper then, and as we will see throughout the following sections, is that Turing's study of human intelligence places itself directly under this normalist paradigm.
\end{hide}

The second normative interpretation of the average is when the normal is only the \emph{mediocre} this time, potentially in need of improvement. 
It is the interpretation typically associated with the British eugenist Francis Galton and with his \emph{Hereditary Genius} in particular~\citep{galton1869}.
Under this normative interpretation\textemdash{}and unlike with that of Quetelet\textemdash{}\emph{some}~(though not all) deviations from the normal/average can prove desirable and be explicitly valued and praised~\citep[see][especially ch.3, for an introduction to Galton's philosophy]{mackenzie1981}.\footnote{
    ``Genius'' in the words of Galton implies two things: \emph{exceptionality} and \emph{innateness}.
    The word expresses ``an ability that [is] exceptionally high, and at the same time inborn''~\citep[p.viii]{galton1869}.
    The genius for Galton is still abnormal\textemdash{}as it characterizes a specific deviation from the~(statistical) norm\textemdash{}but it is abnormal ``good'' this time as it constitutes a ``desirable'' atypicality.
}
It is the normative interpretation that we find behind the modern conception of intelligence underlying psychometrics tests such as IQ tests, which remain fundamentally grounded in the Galtonian figure of the genius~\citep[see][for an overview]{brody2000}.
Galton ultimately came up with the \emph{cumulative} normal curve to better represent the idea that the extreme right hand side of the normal curve, specifically deviating from the mediocre average, was in some circumstances the desired target.
At the far right of the cumulative normal curve in Figure~\ref{fig:normal:cumulative}, you precisely find yourself ``on top'' of everybody else while at the average dashed line at the center of the curve, you are ``just \emph{average}'', so to speak, since $50\%$ of the data points will be below you and $50\%$ above you. 
\begin{hide}
    Where Quetelet emphasized averaging, Galton emphasized \emph{ranking}~\citep[p.33]{davis1995} and in particular the ranking of human behavior with respect to a different normative point of reference than the one normalism was traditionally used to. 
    Where normalism indeed mostly (if not exclusively) focused on normal/average behavior (more often median than mean in fact under Galton's conception) ``intelligence'' in the mind of Galton came to refer to \emph{exceptional} rather than \emph{normal} human behavior. 
\end{hide}

Distinct as they may be, those two normative interpretations of the normal/average remain nonetheless fundamentally intertwined, as the exceptional is always defined in relation to the~(statistically) normal and as a specific deviation from it.
What is more, even when the normal/average does not constitute a normative \emph{ideal}\textemdash{}as with the Galtonian interpretation of the average as only mediocre\textemdash{}it remains at the very least an \emph{idealization} since nobody is ever \emph{perfectly} normal.
Nobody measures \emph{exactly} the ``average height'' for instance, especially not when that average height is computed with infinite precision.
Under the statistical interpretation of normality, ``being normal'' is thus more often than not a matter of remaining under an arbitrary threshold and of being ``close enough'' from an idealized value to be considered falling within its scope.

Within the field of artificial intelligence, then, this idea of ``normative ideal'' takes on a particular meaning as it questions the type of human behavior that will be modeled through the development of artificial minds: normal/average or exceptional? 
The normative, as we have seen, specifies what is ``correct'' in a sense and so the specific question at stake is that of what constitutes ``correct behavior'' as far as machines are concerned. 
Will machines target normal/average or exceptional human intelligence? 
From those two distinct normative ideals will arise two distinct scientific projects for the development of artificial minds: \emph{artificial intelligence} on one side and what I propose to call \emph{artificial smartness} on the other~(see Table~\ref{tab:intsmart:artificial}).
\begin{table}[htpb]
    \centering 
        \resizebox{.7\linewidth}{!}{
            \begin{tabular}{cc}
                \toprule
                \multicolumn{1}{c}{Artificial model} & \multicolumn{1}{c}{Target intelligence} \\
                \midrule
                Artificial intelligence & Normal/Average human intelligence \\
                Artificial smartness & Exceptional human intelligence \\
            \end{tabular}
        }
    \caption{Artificial models of human intelligence: the intelligence/smartness dichotomy.}
    \label{tab:intsmart:artificial}
\end{table}
My motivation for using the term ``artificial intelligence'' to refer to the project of modeling normal intelligence is rooted in the fact that Turing himself explicitly told us that he aimed at modeling the normal/average/mediocre mind of normal people rather than the exceptional mind of geniuses. 
As he is reported to have said to his colleagues at the Bell Laboratories during his 1943 stay indeed: 

\begin{quote}
    I'm not interested in developing a \emph{powerful} brain. All I'm after is just a \emph{mediocre} brain, something like the President of the American Telephone and Telegraph Company.~\citep[Alan Turing, as quoted in][p.316]{hodges1983} 
\end{quote}
In the above quote, Turing makes clear that he places himself directly under the Galtonian normative interpretation of the normal as mediocre while at the same time distancing himself from it by shifting the target of inquiry from the exceptional to the normal.
In doing so, he brings back intelligence into normalism in full and to its dedicated object object of study: the normal.
Under Turing's approach, the target of artificial intelligence thus becomes the normal/average/typical intelligence characteristic of a normal/average/typical human mind.  
%

%

\subsection{From intelligence to normal human behavior}
\label{sec:def:intelligence}

\begin{hide}
    Section~\ref{sec:def:intelligence} introduces Turing's conception of intelligence in more details, tracing back its origins to the notions of \emph{intellect}, \emph{understanding} and the cartesian concept of \emph{reason}.
    It proposes to revisit the distinction between \emph{intelligence} and \emph{intelligent behavior} through the fundamental contribution of \emph{development}, emphasizing that Turing's conception of intelligence as a faculty implies that intelligence is always but a potential that can only become behaviorally manifest through gradual exposure to environmental stimuli.
    After comparing Turing's approach to the faculty of intelligence to Chomsky's approach to the faculty of language, it concludes that the Turing test remains fundamentally normalist in that it does not aim to characterize the boundaries of human cognition\textemdash{}determining all that can and cannot be human\textemdash{}but only \emph{one} of its possible manifestation: the intelligent behavior associated with a particular community of human beings. 
    Such considerations explain, in turn, why the Turing test cannot be taken to provide an operational characterization of intelligence\textemdash{}i.e., providing both a \emph{necessary} and \emph{sufficient} condition for the possession of human intelligence. 
    Considering otherwise would be like saying that manifesting a particular type of developmentally-situated linguistic behavior\textemdash{}say, speaking English\textemdash{}provides both a necessary and sufficient condition for the possession of the faculty of language, which everyone will agree is absurd~(at least for the necessary part).
\end{hide}

The first thing to bare in mind when taking about intelligence as currently and conventionally understood by psychologists is that it is ``a brashly modern notion''~\citep[p.21]{daston1992}.
Before Galton operated a radical shift in the conception of intelligence leading to our modern and everyday understanding of the term, intelligence was primarily conceived as a universal faculty and as something that one either \emph{has} or \emph{does not} have rather than something that one can have \emph{more or less} of~\citep[see][and references therein for overviews]{malabou2019,cave2020}.
As~\citet[p.45]{gonzalez1979} makes clear: ``In the eighteenth and in most of the nineteenth centuries it was thought that humans were endowed by nature with similar, if not identical, powers of the mind''.
Galton thus shifted the notion of intelligence from the universal to the singular and from an essential characteristic of our human nature to a singular attribute belonging only~(in its highest form at least) to a ``happy few''~\citep[p.20]{malabou2019}.
He turned what was so far considered a shared attribute of all human beings into a mechanism that could be used to \emph{compare}\textemdash{}and ultimately \emph{rank}\textemdash{}each and every one of us with respect to some specific normative ideal or normative point of reference.
This gave us the conception of intelligence that we have today\textemdash{}a conception that is fundamentally articulated around~(individual) variability rather than universality, as is typically illustrated by the 1996 report of the American Psychological Association~(APA) on \emph{Intelligence: Knowns and Unknowns}, which begins as follows:

\begin{quote}
    Individuals differ from one another in their ability to understand complex ideas, to adapt effectively to the environment, to learn from experience, to engage in various forms of reasoning, to overcome obstacles by taking thought.~\citep[p.77]{neisseretal1996}
\end{quote}

The conception of intelligence understood as a universal endowment of the human species has a long history in North Atlantic philosophy, tracing back to the notions of \emph{intellect}, \emph{understanding} and the Greek concept of \emph{Noûs}~\citep[see][ch.1]{malabou2019}.\footnote{
    See also~\citep{goldstein2015} for an etymological overview of the term ``intelligence''.
}
%
%
Turing, as it so appears, places himself directly within this longstanding tradition. 
In his seminal \citeyear{turing1950} paper indeed, he introduces the concept of machine intelligence through the question ``Can machines think?'', showing thereby that he uses the two terms \emph{intelligence} and \emph{thinking} interchangeably.
Such a conception of intelligence, however, remains fundamentally at odds with our everyday use of the term, as~\citet{shieber2004a} rightly stresses in his introduction to the Turing test: 

\begin{quote}
    Turing used the terms ``think'' and ``be intelligent'' as if they were synonyms, as one can tell by a simple comparison of his article's title and first sentence. In common usage, the two often mean quite distinct things. When I say that my son is intelligent, I usually mean something beyond the fact that he is capable of thought. However, I and many authors follow Turing's practice, taking the notion of ``being intelligent'' under which it means ``being capable of thought'', rather than ``being smart''.~\citep[footnote 2, p.6]{shieber2004a}
\end{quote}
\citet{block1981} argues along similar lines when discussing Turing's original question:

\begin{quote}
    Note that the sense of ``intelligent'' deployed here\textemdash{}and generally in discussion of the Turing Test [footnote omitted]\textemdash{}is \emph{not} the sense in which we speak of one person being more intelligent than another. ``Intelligence'' in the sense deployed here means something like the possession of thought or reason.~\citep[p.8]{block1981} 
\end{quote}
Intelligence, under the framing of the Turing test, is primarily approached as a human \emph{faculty}\textemdash{}a species-level~(maybe even species-\emph{specific}) ability to ``think'' that one possesses by virtue of being human\textemdash{}and this ``faculty of thinking'' appears to directly echo the ``faculty of understanding'' that characterizes the historical~(i.e. pre-Galtonian) conception of intelligence~\citep[see][p.3]{malabou2019}.

Examples illustrating such considerations do not stop at Turing's framing of the problem, however. 
Some have also pointed out that Turing's conception of the mind~\citep[underlying notably his concept of ``child machine'', see][pp.455\textendash460, for details]{turing1950} bears strong resemblance to Locke's~(\citeyear{locke1690b}) \emph{tabula rasa} theory~\citep[e.g.][]{mays2001}. 
%
Such a lockean conception of the mind as a ``blank slate'', however, entails the general equality of the mind between individuals~\citep[p.45]{gonzalez1979} which echoes once again the concept of ``faculty'' as a universal property of all human beings.
%
Others, such as~\citet{shieber2004},\footnote{
    See also~\citep{abramson2011}.
} typically trace back the origins of the Turing test to~\citet{descartes1637} and to his concept of \emph{reason}~(understood as a \emph{rational soul}) which is not surprising given that the concept of intelligence as a faculty ``played the same role as did reason during the Enlightenment''~\citep[p.1]{malabou2019}.\footnote{
    Saying that Turing's conception of intelligence as a universal property of all human beings places itself directly in the footsteps of a certain cartesian tradition does not mean that Turing adheres to Descartes' argument that ``thinking'' is impossible without reason, however, as he makes clear in his answer to what he calls ``The Theological Objection''~\citep[see][p.443]{turing1950}.
}

\begin{hide}
    \begin{quote}
        $[\ldots]$ if any such machines resembled us in body and imitated our actions insofar as this was practically possible, we should still have two very certain means of recognizing that they were not, for all that, real human beings. [footnote omitted]
        The first is that they would never be able to use words or other signs by composing them as we do to declare our thoughts to others. For we can well conceive of a machine made in such a way that it emits words, 
        $[\ldots]$
        but it is not conceivable that it should put these words in different orders to correspond to the meaning of things said in its presence, as even the most dull-witted of men can do. \citep[p.46]{descartes1637}
    \end{quote}
    In the above passage, Descartes points at what constitutes two core assumptions of the Turing test: first, that linguistic behavior is a fundamental expression of our human nature and second, that this human nature is actually distinct from what we would today characterize as ``intelligence'', as it is manifested even by ``the most dull-witted of men''. 
    As he further insists:

    \begin{quote}
        $[\ldots]$ it is a very remarkable fact that there are no men so dull-witted and stupid, not even madmen, that they are incapable of stringing together different words, and composing them into utterances, through which they let their thoughts be known; and, conversely, there is no other animal, no matter how perfect and well endowed by birth it may be, that can do anything similar. \citep[p.47]{descartes1637}
    \end{quote}
    Central to the reflection of Descartes above is the question of what constitutes the distinctive feature that separates humans from non-human animals. 
    To him, that distinctive feature is \emph{reason}~(understood as a \emph{rational soul}) and it is in fact not surprising to see the origins of the Turing test be traced back to Descartes since, as \citet[p.1]{malabou2019} tells us, the concept of intelligence as a faculty ``played the same role as did reason during the Enlightenment''.
\end{hide}


In sum, intelligence for Turing is a fundamental human ability, a property of our universally shared human nature and \emph{not}, contrary to what our modern conception of the term suggests, a normative point of reference that serves to rank us with respect to one-another.
This is precisely why some authors such as~\citet[p.8]{fostel1993} have been able to argue that the Turing test ``is testing humanity, not intelligence''.

Now of course, approaching intelligence as a faculty and a universal endowment of the human species does not mean that it is necessarily readily available to every human being from birth. 
Like other biological faculties\textemdash{}such as \emph{language} or \emph{vision}\textemdash{}it can be conceived as an ``organ'' that must grow and develop, notably in interaction with its environment and through exposure to external stimuli, before it can turn into an effective ability manifesting itself through observable \emph{behavior}.
%
Until then, it would remain but a mere ``potential'', as Turing himself makes clear in his earlier writings:

\begin{quote}
    \phantomsection\label{quote:turing:potential}
    $[\ldots]$ the potentialities of the human intelligence can only be realised if suitable education is provided.~\citep[pp.431\textendash432]{turing1948}
\end{quote}
Between \emph{intelligence} and \emph{intelligent behavior} thus lies the fundamental contribution of \emph{development}, which is why the two cannot really be considered perfectly equivalent, even for Turing.

However, if Turing renews with a longstanding tradition in the conceptualization of intelligence, he also distances himself from it in a significant way.
For what constitutes the proper target of his scientific inquiry is not this faculty of intelligence \emph{per se} but only its overt manifestation: intelligent \emph{behavior}.
\begin{hide}
    To better understand this point, it may be useful here to compare the work of Turing to that of Chomsky\textemdash{}itself dedicated to another fundamental human faculty: the faculty of \emph{language}~\citep[see][for a brief overview]{chomsky2017b}.
    To characterize the content of the language faculty is, for Chomsky, to delineate the boundaries of human cognition\textemdash{}to understand, in some sense, what \emph{can} and \emph{cannot be} human. 
    The ultimate goal of linguistics under the Chomskyan approach is thus ``To define the class of possible human languages''~\citep[p.1]{moro2016}.
    The concept of \emph{possible} here opposes another fundamental epistemological concept: the \emph{probable} and, through it, the concept of statistical normality.
    Within the Chomskyan approach to linguistics indeed, statistically prevalent linguistic behaviors play no particular role in investigating the boundaries of human cognition: nothing makes ``English'' or ``Mandarin'' more valid objects of study than ``Quechua'' or ``Breton'', for instance, as exhibiting a higher number of speakers does not make a particular language a more relevant object of study when the ultimate goal is to determine the boundaries of human cognition.
    %
    %
    The statistical prevalence of a particular language or sets of linguistic behaviors is thus irrelevant to the Chomskyan approach to linguistics, as its paradigm is fundamentally dedicated to what \emph{can} be human and not to what is \emph{likely} to be human.\footnote{
        If anything, the Chomskyan paradigm actually requires us to pay particular attention to the atypical and the~(statistically) abnormal since, ``unlikely'' as it may be, it remains nonetheless a possible manifestation of what can be human.
    }
    It is in that sense that we can say that the Chomskyan approach to language is ``anti-normalist'': the concept of statistical normality plays little to no role in its epistemology and does not constitute its target of scientific inquiry.\footnote{
        Which is why, in passing, Chomsky and colleagues have been so critical of the use of probabilistic models in linguistics~\citep[see][]{chomsky1957,chomsky1968,chomskyetal2023,moroetal2023,bolhuisetal2024}.
        When Chomsky tells us that ``probabilistic models give no particular insight into some of the basic problems of syntactic structure''~\citep[p.17]{chomsky1957} or that ``it must be recognized that the notion of `probability of a sentence' is an entirely useless one, under any known interpretation of this term''~\citep[p.57]{chomsky1968}, his argument is not\textemdash{}contrary to what some computational linguists may have interpreted~\citep[e.g.][]{pereira2000,norvig2012}\textemdash{}a mere technical point regarding the capabilities of statistical language models to capture nuances of grammaticality~(such as the difference between the grammaticality of the sentence \emph{colorless green ideas sleep furiously} and the agrammaticality of the sentence \emph{furiously sleep ideals green colorless}). 
        What is at stake is a central epistemological argument regarding the role that the probable~(and statistical normality in general) is supposed to play within the language sciences.
        As far as Chomsky's approach is concerned, the matter is clear: the critical concept is that of the \emph{possible} and not the probable.
        Which is precisely why his recent critique of large language models~(LLMs) such as ChatGPT focuses on the fact that ``they are incapable of distinguishing the possible from the impossible''~\citep{chomskyetal2023}. 
        The argument is taken on by~\citet[p.84]{moroetal2023} who tell us that ``the distinction between possible versus impossible languages cannot be formulated by definition for LLM'' as well as by~\citet[p.489]{bolhuisetal2024} who argue that ``LLMs can produce `impossible' languages, not generated by the principles governing all known human languages, just as well as~(if not better than) natural language output, and cannot distinguish between them''.
        Else, the fact that Chomskyan linguistics constitutes a fundamental divorce from the normalist paradigm\textemdash{}as it attempts to bring back human nature in place of normality as its central organizing concept\textemdash{}is made explicit by~\citet[pp.63\textendash65]{hacking1996b} and by Chomsky himself~(though somewhat less explicitly) in his \emph{Cartesian Linguistics}~\citep{chomsky1966}. 
    } 
\end{hide}
Indeed, the Turing test does not aim at delineating the boundaries of human cognition or at characterizing \emph{all} forms of intelligent behavior for that matter.
%
It does not actually provide an ``operational'' definition of intelligence as there is no equivalence between possessing intelligence and successfully passing the Turing test~\citep{moor1976,moor2001,copeland2000}.
Turing makes that clear in his original paper when he tells us that: 

\begin{quote}
    May not machines carry out something which ought to be described as thinking but which is very different from what a man does? This objection is a very strong one, but at least we can say that if, nevertheless, a machine can be constructed to play the imitation game satisfactorily, we need not be troubled by this objection. \citep[p.435]{turing1950}
\end{quote}
Failing to pass the Turing test thus does \emph{not} rule out something~(or someone) as being intelligent.
At most, what we can is that the Turing test intends to provide a \emph{sufficient} condition for the possession of intelligence, but certainly not a \emph{necessary} one.
What makes Turing's approach remain fundamentally normalist in the end 
is that it does not target the boundaries of the faculty of intelligence but only its typical manifestation\textemdash{}or at least \emph{one} of its typical manifestation.
%
%
For indeed, note how, in the above quote, \citet[pp.431\textendash432]{turing1948} employs the term ``education'' rather than ``development'' to discuss the transition from intelligence to intelligent behavior\textemdash{}using a term that evokes a rather socio-culturally and socio-historically situated practice in comparison, as it seems much less straightforward to talk about ``human education'' than to talk about ``human development''. 
Turing's terminological choice thereby suggests, even if only implicitly, that what we take to be ``intelligent behavior'' may depend on the ``education'' we have received and so, ultimately, on the particular group of humans we belong to.\footnote{
    And indeed, many have argued that the normative point of reference we use to rank intelligent behaviors\textemdash{}or what constitutes the \emph{set} of intelligent behaviors to begin with\textemdash{}is both socio-culturally and socio-historically situated~\citep[see][for overviews]{vernon1965,vernon1969,berry1972,berry1980,sternberg1984,sternberg1985,sternberg2004}.
}
This explains why the Turing test can be considered a test of ``culturally-oriented human intelligence'' that ``could be passed only by things that have experienced the world as we have experienced it''~\citep[p.53]{french1990}.
\begin{hide}
    The argument is not restricted to intelligence: we always develop all sorts of group-specific developmental features depending on our environment we grow up in. 
    For instance, we all acquire sets of group-specific linguistic behaviors depending on the particular linguistic stimuli we are exposed to~(especially during childhood). 
    Thus, we all end up speaking or signing a particular ``language'', be it English, Turkish, Swahili, etc., that is always both socio-culturally and socio-historically situated, despite our purportedly sharing a common language faculty.\footnote{
        The argument extends to other fundamental human faculties such as \emph{bipedalism}, for instance, for which upright posture and bipedal gait have been considered ``outstanding features of human nature''~\citep[p.687]{hewes1961}.
        Anthropologists have indeed long shown that the human bipedal striding gait is affected by socio-cultural factors so that ``The striding gait would be better viewed as a biologically based cultural trait that predominates as a locomotor form in our own society but one that should not be applied as such to other groups in an ethnocentric way''~\citep[p.551]{devine1985}.
        As~\citet[p.553]{devine1985} elaborates: ``Everyone is aware of subtle differences in gait that exist between individuals. One of the first experiences of early visitors to non-Western societies was an immediate realization that the manner of walking of the people they were encountering was quite dissimilar from their own''.
    }
\end{hide}
The ambition of the Turing test is therefore quite modest, all things considered, as it does not aim to model \emph{all} possible normal human behaviors or even \emph{the} normal human behavior~(in the sense of the most statistically prevalent behavior within the human population as a whole) but only \emph{a} possible normal human behavior\textemdash{}the average/typical behavior of a particular community of human beings.  

\section{Normality against exceptionality}
\label{sec:smart}

\subsection{The Turing test is a test of normal intelligence}
\label{sec:arg:normint}

\begin{hide}
    Section~\ref{sec:smart} revisits the Turing test through the normal/exceptional dichotomy. 
    Section~\ref{sec:smart:error} introduces the fundamental distinction between \emph{intelligence} and \emph{smartness} focusing on the opposition between intelligent behavior understood as normal/average human behavior on one side and smart behavior understood as exceptional human behavior on the other.
    The notion of smart behavior is further refined as a \emph{normative ideal} of \emph{correct} in addition to \emph{exceptional} human behavior, which enables two things. 
    First, framing smart behavior as a \emph{normative ideal} of human behavior explains how smartness characterizes an \emph{ideal} type of human behavior from which most if not all real/normal people deviate. 
    Second, framing smart behavior as a normative ideal of \emph{correct} human behavior explains why the Turing test, as a test of normal/average human behavior, specifically requires machines to \emph{make mistakes} and deviate from the normative ideal of smart behavior in the way normal people would.
    The intelligence/smartness dichotomy is then illustrated through the example of \emph{spelling} as perfect spelling is taken\textemdash{}in English at least\textemdash{}to constitute a prototypical example of a normative ideal on human behavior that normal people ought to abide to but never quite manage to in practice.
    The example provides a concrete way to illustrate the fundamental difference between the two scientific projects that constitute \emph{artificial smartness} on one side and \emph{artificial intelligence} on the other: the former being dedicated to the modeling of correct and exceptional human behavior in general and perfect/correct spelling in particular while the latter is dedicated\textemdash{}for Turing at least\textemdash{}to the modeling of normal/average human behavior in general and imperfect/normal/average spelling in particular.
\end{hide}

As we understand from the previous section, it is not so much Turing's conception of intelligence as a faculty that makes it remain at odds with our everyday use of the term\textemdash{}as it does not really constitute the target of his scientific inquiry\textemdash{}but his conception of intelligent behavior as \emph{normal} rather than \emph{exceptional} human behavior.
To make the matter clearer, what I propose in this section is to reframe the opposition between normal and exceptional human behavior as the opposition between intelligent and smart behavior so as to better introduce the critical distinction I wish to make between artificial intelligence and artificial smartness. 
%
My core contribution in this section is that artificial intelligence and artificial smartness correspond to two different scientific projects for the construction of artificial minds, each targeting a different normative ideal of human behavior: the normal on one side and the exceptional on the other~(see Table~\ref{tab:intsmart:overview}).  
\begin{table}[htpb]
    \centering 
        \resizebox{\linewidth}{!}{
            \begin{tabular}{ccc}
                \toprule
                \multicolumn{1}{c}{Artificial model} & \multicolumn{1}{c}{Target behavior} & \multicolumn{1}{c}{Human reference}  \\
                \midrule
                Artificial intelligence & Intelligent behavior & Normal/Average human behavior \\
                Artificial smartness & Smart behavior & Exceptional human behavior \\
            \end{tabular}
        }
    \caption{Intelligence against smartness, or normality against exceptionality}
    \label{tab:intsmart:overview}
\end{table}

The tension between~(artificial) intelligence and~(artificial) smartness is present throughout the literature on artificial minds, albeit under different forms and using different terminologies.
Harnad, for instance, approaches the normal/exceptional dichotomy through the real/ideal one\textemdash{}emphasizing thereby how smart behavior characterizes a normative \emph{ideal} of human behavior.
He stresses that, to successfully pass the Turing test, machines need to be able to do ``many (perhaps most, perhaps all) of the things a real person can do''~\citep[p.43]{harnad1991}.
He thus opposes ``real'' people who display normal/typical/average human behavior to geniuses like Einstein who, if no less ``real'' than normal people, still constitute significant deviations from the normal/average and, as such, cannot constitute the target of machines aiming to pass a Turing test which requires them to be indistinguishable from normal people: 

\begin{quote}
    As to Einstein well, it seems a rather high standard to hold computers to, considering that almost every one of us would probably likewise fail resoundingly to meet it.~\citep[p.43]{harnad1991}
\end{quote}
Others, such as Russell and Norvig, mobilize a concept of \emph{reason} or \emph{rationality}, loosely defined as the ability to ``do the right thing''~\citep[p.31]{russellandnorvig2020} and which emphasizes this time the \emph{normative} aspect of an ideal of smart behavior from which most people deviate:

\begin{quote}
    We are not suggesting that humans are ``irrational'' in the dictionary sense of ``deprived of normal mental clarity.'' 
    We are merely conceding that human decisions are not always mathematically perfect.~\citep[p.31]{russellandnorvig2020}
\end{quote}
In practice indeed, real people ``make mistakes'', and that is precisely what makes them humans.\footnote{
    \citet[p.397]{pinsky1951} notably writes that humans are ``unique [in the universe] by virtue of the ability to \emph{mis}use the faculty of reason''.
}

Such considerations underline two things. 
First, that smart behavior constitutes a normative ideal of \emph{correct}~(if not \emph{perfect}) human behavior, in addition to being a normative ideal of exceptional human behavior from which normal/average people deviate. 
Second, that intelligent behavior for Turing, as a particular deviation from this normative ideal of correct and exceptional human behavior, specifically targets the imperfections of our human nature. 
Turing makes clear indeed in earlier writings that ``If a machine is expected to be infallible, it cannot also be intelligent''~\citep[p.105]{turing1947b}.
In his original \emph{Mind} paper, he explicitly tells us that a machine would have to ``deliberately introduce mistakes in a manner calculated to confuse the interrogator'' so as to be able to pass his test~\citep[p.448]{turing1950}.
To successfully pass the Turing test, a machine thus needs to be able to ``make mistakes'' so as to behave in an ``incorrect'' or ``imperfect'' way\textemdash{}just like normal people would.


The best way to understand the dichotomy between intelligent and smart behavior is through the example of \emph{spelling}~(see Table~\ref{tab:intsmart:example}) as it constitutes\textemdash{}in English at least\textemdash{}a typical example of a normative ideal of correct behavior from which most if not all people deviate.\footnote{
    ``Mistakes are a fact of life'' write Lunsford and Lunsford in their review of first year college students writing in the USA, for which they report a typical error rate of 2 to 3 mistakes per 100 word~\citep[see][Table 8]{lunsfordandlunsford2008}.
    Note that their concept of ``error'' includes notions such as ``vague pronoun reference'' or ``unnecessary comma'' which probably extends much beyond what is relevant to our present discussion, but that at the same time their concept of ``spelling error'' excludes notions such as ``missing word/comma/hyphen/capitalization'' which I would definitely include in our concept of ``spelling mistake''~\citep[see][and notably Table~4 and Table~7, for details]{lunsfordandlunsford2008}.
    Flor and colleagues otherwise report that only 10.7\% of essays from native speakers of English in the TOEFL or GRE exams contain no spelling mistake~\citep[see][Table 3]{floretal2015}.
} 
\begin{table}[htpb]
    \centering 
        \resizebox{\linewidth}{!}{
            \begin{tabular}{ccc}
                \toprule
                \multicolumn{1}{c}{Target behavior} & \multicolumn{1}{c}{Human reference} & \multicolumn{1}{c}{Example}  \\
                \midrule
                Intelligent behavior & Normal/Average human behavior & Imperfect~(normal/average) spelling \\
                Smart behavior & Exceptional human behavior & Perfect spelling \\
            \end{tabular}
        }
    \caption{Intelligence against smartness: the example of spelling.}
    \label{tab:intsmart:example}
\end{table}
Even the best speller, I would argue, is bound to make a spelling mistake at some point, whether it is because of temporary fatigue, distraction, memory lapse, etc., and even though they may otherwise know perfectly well the correct form of the expression at hand.
In fact, Turing explicitly said in a 1952 BBC interview that, in order to successfully pass his test, ``the machine would be permitted all sorts of tricks so as to appear more man-like, such as waiting a bit before giving the answer, or making spelling mistakes''~\citep[p.495]{turingetal1952}.
And indeed, practical Turing tests carried out in 2008 at the University of Reading~\citep{shahandwarwick2010a,shahandwarwick2010c}, or in 2012 at Bletchley Park~\citep{warwickandshah2014a,warwickandshah2016c} have since shown that the production of perfect spelling is often precisely what makes machines detectable by human judges. 
Warwick and Shah note, for instance, that, on one occasion, ``the judge did correctly identify the human entity as there were a lot of spelling mistakes in their discourse and the conversation was quite stilted'' and conclude from their review of practical Turing tests that ``the occasional spelling mistake seems to add human credibility''~\citep[see][pp.1001 and 1003, respectively]{warwickandshah2016a}.

Acknowledged as it may be by the literature, the distinction between artificial intelligence and artificial smartness is nonetheless usually framed as a distinction between ``useless'' and ``useful'' artificial models, emphasizing thereby the potential limitations of a Turing test dedicated to ``incorrect'' human behavior of some sort. 
Hayes and Ford, for instance, argue in their paper \emph{Turing Test Considered Harmful} that ``Our most useful computer applications~(including AI programs) are often valuable exactly by virtue of their lack of humanity'' and that ``A truly human-like program would be nearly useless''~\citep[p.975]{hayesandford1995}.
They conclude accordingly that the Turing test ``is no longer a useful idea'' and even that ``it is now a burden to the field, damaging its public reputation and its own intellectual coherence''~\citep[p.972]{hayesandford1995}.
Their argument rejoins that of French who would later argue that:

\begin{quote}
    $[\ldots]$ we need to put aside the attempt to build a machine that can flawlessly imitate humans; for example, do we really need to build computers that make spelling mistakes or occasionally add numbers incorrectly, as in Turing's original article~[footnote omitted] in order to fool people into thinking they are human?~\citep[pp.74\textendash75]{french2012}
\end{quote}
The argument is not new: ``who needs a machine that can't type?'' asked \emph{The Economist} already in a~\citeyear{theeconomist1992} article on \emph{Artificial Stupidity} dedicated to the results of the first edition of the \emph{Loebner Prize Competition} held at the Computer Museum in Boston on November 8 1991~\citep[see][for details]{epstein1992}.\footnote{
    Note that it is precisely the results of those experiments that lead Fostel to introduce his humanity/intelligence dichotomy:
    \begin{quotation}
        The recent Loebner trial [\citep{theeconomist1992}], in which a computer program achieved considerable success, perhaps passing [the Turing test] as construed for that trial, demonstrates further that [the Turing test] tests for humanity. 
        The computer program used carefully calculated typing errors to fool [the human judge] into believing it to be human. 
        The capacity to make statistically human-like typing errors is not a convincing step towards intelligence yet it seems to be a major step towards passing [the Turing test]. It made the program more human.~\citep[p.8]{fostel1993}
    \end{quotation}
}

The vocabulary associated with artificial smartness is typically that of \emph{useful} and \emph{profitable} models dedicated to \emph{realistic} tasks and \emph{practical} problems\textemdash{}by opposition to artificial intelligence models aiming at successfully passing the Turing test and which, as such, are considered~(even if only implicitly) useless and unprofitable artifact dedicated to unrealistic and abstract tasks and problems.
Such is the case, for instance, of~\citet[p.64]{bringsjordandgovindarajulu2022} who, in their encyclopedic entry on \emph{Artificial Intelligence}, tell us that ``most AI researchers and developers, in point of fact, are simply concerned with building useful, profitable artifacts, and don't spend much time reflecting upon the kinds of abstract definitions of intelligence explored in this entry~(e.g., What Exactly \emph{is} AI?)''. 
Or of~\citet[p.1836]{russellandnorvig2020} who argue that ``Few AI researchers pay attention to the Turing test, preferring to concentrate on their systems' performance on practical tasks, rather than the ability to imitate humans''. 
In fact, Turing himself wrote, in a letter to the cyberneticist W. Ross Ashby: ``I am more interested in the possibility of producing models of the action of the brain than in the practical applications to computing''~\citep[see][p.374]{turing1947}.
%
However, framing the distinction between artificial intelligence and artificial smartness in terms of absolute usefulness is somewhat misleading, as nothing is ever intrinsically ``useful'' in and of itself.
In effect, nothing makes artificial smartness \emph{inherently} more ``useful'' or ``profitable'' than artificial intelligence\textemdash{}everything depends on the target application and on its context of use.\footnote{
    In fact, \citet[p.8]{fostel1993} would even argue that artificial smartness was useless as ``No one is going to want a vacuum cleaner that speaks better English than they do''. 
}
This is why I propose to introduce those two notions as simply referring to two different scientific projects dedicated to the modeling of two different normative ideals of human behavior~(see Figure~\ref{fig:artificial:intelligence} and Figure~\ref{fig:artificial:smartness}).
\begin{figure}[htbp]
    \begin{center}
        \begin{minipage}{0.44\textwidth}
            \centering
            \resizebox{\linewidth}{!}{
                \begin{tikzpicture}
                    \begin{axis}[
                        ticks=none,
                        samples=100,
                        yticklabel=\empty,
                        xticklabel=\empty
                        ]
                        \addplot [thick] {gauss(0,1.5)};
                        \draw [thick, blue] (0,0 |- current axis.south) -- (0,0 |- current axis.north);
                    \end{axis}
                \end{tikzpicture}
            }
            \caption{Artificial intelligence, or the modeling of \emph{normal} human behavior. Target behavior corresponds to perfectly normal/average human behavior marked as a straight blue line at the center of a normal curve modeling a hypothetical distribution over human behavior.}
            \label{fig:artificial:intelligence}
        \end{minipage}
        \hspace{.1\textwidth}
        \begin{minipage}{0.44\textwidth}
            \centering
            \vspace{1.1cm}
            \resizebox{\linewidth}{!}{
                \begin{tikzpicture}
                    \begin{axis}[
                        ticks=none,
                        samples=100,
                        yticklabel=\empty,
                        xticklabel=\empty,
                        legend style={
                            at={(0.5,-0.1)}, 
                            anchor=north
                        }
                        ]
                        \draw [dashed] (0,0 |- current axis.south) -- (0,0 |- current axis.north);
                        \draw [thick, blue] (5,0 |- current axis.south) -- (5,0 |- current axis.north);
                        \addplot [thick] {normcdf(x,0,1.5)};
                    \end{axis}
                \end{tikzpicture}
            }
            \caption{Artificial smartness, or the modeling of \emph{exceptional} human behavior. Target behavior corresponds to exceptional human behavior marked as a straight blue line at the far right of a cumulative normal curve modeling a hypothetical distribution over human behavior. 
            Note how the straight blue line denoting exceptional human behavior deviates from the normal/average dashed line at the center of the curve.}
            \label{fig:artificial:smartness}
        \end{minipage}
    \end{center}
    \label{fig:int:vs:smart}
\end{figure}

\subsection{Why ChatGPT is unlikely to pass the Turing test}
\label{sec:arg:chatgpt}
\begin{hide}
    x. why ChatGPT won't pass the TUring test 
    
    x. why it is not an ontological argument 

    x. why not any mistake goes 

    x. why it is a bad model of normal mistakes anyway (at least at the moment)
\end{hide}
\begin{hide}
    With those distinctions in mind, Section~\ref{sec:smart:chatgpt} introduces the core implications of the normalist reading of the Turing test for large language models such as ChatGPT.
    Its main argument is that such models are unlikely to pass the Turing test as they are models of artificial smartness rather than artificial intelligence\textemdash{}targeting exceptional rather than normal human intelligence. 
    The argument however, is analytical and not ontological. 
    That is, its point is not to claim that such models could \emph{never} be made so as to successfully pass the Turing test but only that, for the practical applications those models are usually being put to use, it makes more sense for them to model smart/exceptional/correct human behavior rather than intelligent/normal/incorrect human behavior. 
    The argument is plain: \emph{who needs machines that make mistakes?}
    That being said, if nothing prevents us from training those very same models on new data so as to learn how to ``make mistakes'', what is crucial to understand is that \emph{not any mistakes goes} as far as the Turing test is concerned: to successfully pass the Turing test, those models would need to produce \emph{normal/average mistakes} as normal people would. 
    And determining the nature of those normal mistakes\textemdash{}and creating the corresponding datasets along with it\textemdash{}is quite a scientific challenge. 
    It is, in fact, the very challenge that \emph{artificial intelligence} as originally conceived by Turing was supposed to tackle.
\end{hide}

With those distinction in mind, we can now better make sense of recent results arguing that ChatGPT falls short of passing the Turing test~\citep[e.g.][]{restrepoEchavarria2025}.
They stand in the long tradition of studies which have consistently argued that machines exhibiting exceptional human behavior would not be capable of passing the Turing test.
Back in 1991, for instance, and after a Loebner Prize Competition which saw a human participant being repeatedly misclassified as a computer due to her extensive knowledge of Shakespeare~\citep[p.88]{epstein1992},\footnote{
    See also~\citep[p.72]{shieber1994}.
} Fostel would argue that:

\begin{quote}
    Suppose [an alien] robot had scanned and absorbed all of Earth's libraries, media, daily conversations and so on. Would the robot be able to pass [the Turing test]? I think not. 
    $[\ldots]$
    the Robot will be so different from any available human~$[\ldots]$ that this failure in discrimination by [a human judge] would be unlikely. 
    A nearly omniscient agent~$[\ldots]$ would dearly not be like any human~$[\ldots]$ Even if the alien robot is extremely capable, and superior to any human in intellectual endeavors, it will fail [the Turing test].~\citep[p.8]{fostel1993}
\end{quote} 
Fastforwarding to 2025 and Restrepo Echavarría tells us specifically that ChatGPT~(in its GPT-4 version at least) fails to pass Turing test because ``its intellectual abilities surpass those of individual humans'' and because it simply proves ``too smart'' to be human~\citep[see][pp.1 and 5, respectively]{restrepoEchavarria2025}.
In short, ChatGPT fails to pass the Turing test because it is a model of artificial smartness rather than artificial intelligence \emph{per se}~\citep[that is, in the original sense of][]{turing1950}.\footnote{
    Which is precisely how we should interpret the fact that it is reported to score higher than 99.9\% of the population of Verbal IQ tests~\citep{roivainen2023} or that it outperforms most average performance on academic and professional exams~\citep[see][Table 1, for details]{openai2023a}.
    As a model of artificial smartness, ChatGPT targets exceptional human intelligence located at the far right of the normal curve as far as those exam performance are concerned. 
}

Does that mean that ChatGPT could \emph{never} be made so as to successfully pass the Turing test? 
Of course not. 
\begin{hide}
    The point is not to say that there are \emph{intrinsic} properties in the architecture of language models that make them unable to pass the Turing test.
    The argument carried out here is not ontological but analytical. 
    The point is merely to observe that, for the practical applications models such as ChatGPT are currently being put to use, it makes more sense for them to model exceptional rather than normal human behavior\textemdash{}hence their being models of artificial smartness rather than artificial intelligence by default.
    Be that as it may, we have to understand what it takes to successfully pass the Turing test. 
\end{hide}
But we have to understand what it would take to do so and notably that it would take more than mere ``re-programming''~\citep[p.468]{pinarSayginetal2000} or the introduction of a trivial ``mistake module''~\citep[p.3]{lacurts2011}. 
For contrary to a commonly held belief, simply adding ``random'' mistakes to machines output will not do.
To successfully pass the Turing test, machines need not produce \emph{any} mistakes but only \emph{normal/average} mistakes\textemdash{}the mistakes that normal people would make.
As far as ChatGPT is concerned, we do not seem to be quite there yet, for when specifically asked to ``make mistakes'', it still proves incapable of producing \emph{normal} mistakes~\citep[see][\S3, for details]{restrepoEchavarria2025}.
This whole discussion underlies both the depth and complexity of the Turing test.
As~\citet[p.3]{lacurts2011} emphasizes, the fact that we make mistakes as humans ``happens as a result of the internal workings of our brain''.  
Understanding \emph{why} and \emph{how} we make mistakes is therefore a genuine scientific question and an integral part of the cognitive science enterprise. 
What Turing proposed with his test is thus more than just a simple ``game''. 
It is a genuine scientific project characterizing an anthropology of human being dedicated to the understanding of normal/average/typical human behavior.

\begin{hide}
    \citep{lacurts2011}
    \begin{quote}
        \textbf{3.1 The Turing Test Encourages Mistakes}
    \end{quote}
    \begin{quote}
        In [9], Michie criticizes the TT for encouraging machines to make both types of mistakes: “surely one should judge a test as blemished if it obliges candidates to demonstrate intelligence by concealing it.” However, we can resolve this criticism in one of two ways. First, it is possible that mistakes will come about as a result of the machine’s intelligence algorithm. After all, humans generally do not make mistakes intentionally; it \textbf{happens as a result of the internal workings of our brain} (we forget certain facts, etc.). In this case, the TT is fine as it stands. \textbf{Second, if we get to the point of creating machines that could pass the TT but for the fact that they make too few mistakes, the ``mistake module'' would almost surely be trivial to add in.} If it were not, that would be an interesting result in its own right, and certainly those machines should be deemed intelligent and the TT modified. But note that we have to get to the point of almost passing the \emph{current} TT before that is an issue.
    \end{quote}

    the fact that we make mistakes ``happens as a result of the internal workings of our brain'': precisely why the Turing test is a genuine scientific project
\end{hide}

\begin{hide}
    The argument carried out in this section in not an ontological one. 
    \begin{hide}
        That is, the point is not to argue about the intrinsic capabilities of language models but rather to discuss their modeling targets as far as human behavior in concerned. 
    \end{hide}
    The point is not to say that there are intrinsic limitations in the capabilities of language models, constitutive of their architectures for instance. 
    If anything, the limitations of ChatGPT we observe with respect to perfect/imperfect spelling for instance have more to do with the \emph{data} it has been trained on than on its architecture \emph{per se}. 
    Neither is it to say that the fact that ChatGPT is trained on smart rather than intelligent behavior attest of the intrinsic usefulness of artificial smartness models. 
    Once again, the fact that ChatGPT\textemdash{}in its default behavior at least\textemdash{}exhibits a behavior calibrated on perfect spelling rather than normal/average spelling has probably more to do with a socio-economic reality and with the fact that, for most of the applications the system is currently being put to use, it makes more sense for it to exhibit perfect or near-perfect spelling rather than normal/average spelling.\footnote{
        Why would anyone design a language model that deliberately makes spelling mistakes when you know that spelling mistakes can cost you millions in online sales or advertising revenue, affect your online credibility or the perception of the quality of your writing, or that even a single spelling mistake can earn you a million pounds lawsuit in the UK?~\citep[see][and references therein for a comprehensive overview]{panetal2021}.
    }
    In a different context and for specific applications\textemdash{}e.g., create an automated CAPTCHA solver\textemdash{}it may prove more beneficial to target normal rather than exceptional human behavior\textemdash{}in which case the Turing test would prove more relevant than ever.\footnote{
        Note that the distinction between intelligence and smartness is not strictly binary and admits infinite nuance, in the sense that there could exist many use-cases which would require imperfect behavior that is not exactly normal/average~(so something in between intelligent/normal and smart/exceptional human behavior). 
        Think, for instance, of the typical use-case of trying to model the behavior of a ``good'' student, precisely \emph{not} average but not ``perfect'' either as a perfect student would just be too good to be true and therefore too easily detectable by the teacher.
    }
    In fact, this is precisely the point made by Fostel in his original paper who, upon acknowledging that the Turing test targets ``humanity'' rather than ``intelligence''\textemdash{}that is, a form of normal rather than exceptional human behavior\textemdash{}tells us that:

    \begin{quote}
        This does not mean the [Turing test] is useless. 
        It might be a useful marketing tool for 21st century firms producing domestic robots. 
        No one is going to want a vacuum cleaner that speaks better English than they do. 
        Talking alarms in cars have already earned considerable ire by noting minor errors their human occupants make in closing doors. 
        Harnad's [Total Turing test, see~\citep{harnad1992}] is an even better candidate for that marketing function: who wants a baby-sitting robot that makes mom or dad look bad?~\citep[p.8]{fostel1993}
    \end{quote}
\end{hide}

\section{Normality and the statistical type}
\subsection{The Turing test is a statistical test}
\label{sec:arg:stat}

\begin{hide}
    x. the Turing test is a statistical test 

    x. statistics are used to objectivize what is normal

    x. why? to overcome the subjectivity of individual human interrogators 

    x. what is objectivized? Not a universal or statistically prevalent human we
\end{hide}

\begin{quote}
    There is, no doubt, an unsophisticated usage according to which what is normal is what is familiar, and the unfamiliar is feared or condemned as abnormal. But since we are all sophisticated this need not detain us.~\citep[p.221]{dupre1998}
\end{quote}
In her paper deconstructing Turing's conception of intelligence, \citet[p.303]{proudfoot2017b} argues against the standard ``behaviorist'' interpretation of the Turing test by claiming that the test does not evaluate machine behavior but only ``the observer's reaction to the machine''~\citep[see also][]{proudfoot2013,proudfoot2020}.
Her argument rejoins that of~\citet[p.4]{watt1996} which claims that ``The ascription of intelligence [in the Turing test] depends on the observer as well as the behaviour of the system'' and that intelligence for Turing may, like beauty, ``truly be in the eye\textemdash{}or the mind\textemdash{}of the beholder''.
And indeed, Warwick and Shah observe from their practical Turing tests that:

\begin{quote}
    $[\ldots]$ subjectivity plays a big part in attributing `humanness' to another on the basis of responses to chosen questions.~\citep[p.11]{warwickandshah2015}
\end{quote}
Such considerations suggest\textemdash{}even if only implicitly\textemdash{}that the Turing test suffers from a major shortcoming in that it fundamentally relies on the subjectivity of the human interrogator to settle on machine intelligence.
This is what make Hayes and Ford argue, for instance, that:

\begin{quote}
    The imitation game conditions say nothing about the judge, but the success of the game depends crucially on how clever, knowledgeable, and insightful the judge is.~\citep[p.973]{hayesandford1995}
\end{quote}
Yet, the argument neglects a fundamental aspect of the Turing test, namely, that it is a \emph{statistical} test where success is never assessed by a single judge but always by a full \emph{jury}~\citep[see notably][pp.38\textendash39]{guccioneandtamburrini1988}.
Turing himself makes that clear in his original paper when he discusses the possibility to carry out his Imitation Game already:

\begin{quote}
    There are already a number of digital computers in working order, and it may be asked, `Why not try the experiment straight away? It would be easy to satisfy the conditions of the game. A number of interrogators could be used, and statistics compiled to show how often the right identification was given.'~\citep[p.436]{turing1950}
\end{quote}

The best way to understand this critical aspect of the test is to return to Turing's original paper once again and to his famous prediction where he tells us that: 

\begin{quote}
    I believe that in about fifty years' time it will be possible to programme computers, with a storage capacity of about $10^9$, to make them play the imitation game so well that an average interrogator will not have more than 70 per cent, chance of making the right identification after five minutes of questioning.~\citep[p.442]{turing1950}
\end{quote}
The statistical aspect of the test becomes clear from the probabilistic framing of success as a matter of~(70\%) \emph{chance} of making the right guess. 
Incidentally, it also stresses how the Turing test is fundamentally an \emph{indistinguishibility} test and not a \emph{misidentification} test, since this 70\% chance implies that machines do not have to be systematically misidentified as humans to successfully pass the test\textemdash{}though strictly statistically speaking, true indistinguishibility involves a 50\% chance of being identified either as human either as machine.\footnote{
    The fact that Turing suggests a 70\% chance rather than 50\% chance in his original paper can probably be put\textemdash{}alongside the choice of considering only 5min of questioning\textemdash{}on the account of trying to provide a reasonably challenging goal for a task that was back then considered hard enough already. 
    In effect, a baseline of 70\% makes the test slightly ``easier'' than a baseline of 50\%, since with the former the average human interrogator is precisely given ``more chance'' to make the right guess than with the latter.
}

But the statistical aspect of the test also conditions how we ought to interpret this notion of ``average interrogator'' that \citet[p.442]{turing1950} talks about in his above quote.
Most of the time, this notion is understood through the normal/exceptional dichotomy and as referring to a normal/average interrogator understood as non-expert.
Copeland, for instance, tells us that: 

\begin{quote}
    The qualification `average' presumably indicates that the interrogators are not to be computer scientists, psychologists, or others whose knowledge or training is likely to render them more skilled at the task than an average member of the population.~\citep[p.527]{copeland2000}
\end{quote}
This is not surprising given that Turing himself says in his BBC interview that: 

\begin{quote}
    The idea of the test is that the machine has to try and pretend to be a man, by answering questions put to it, and it will only pass if the pretence is reasonably convincing. 
    A considerable proportion of a jury, who should not be expert about machines, must be taken in by the pretence.~\citep[p.495]{turingetal1952}
\end{quote}
Note, however, that \citet[p.442]{turing1950} talks about ``\emph{an} average interrogator''~(singular) in his original paper rather than ``average interrogator\emph{s}''~(plural), suggesting thereby that this notion of ``average'' applies to the \emph{set} of interrogators as a whole rather than to every human interrogator taken individually.
And indeed, it is also possible to interpret this notion of ``average interrogator'' as referring to an \emph{abstraction} made of the aggregate of individual interrogators' judgments\textemdash{}an abstraction corresponding to the standard objectivization practice of the behavioral sciences consisting in averaging individual judgments so as to smooth out intrinsic inter-and intra-subject variability.\footnote{
    The two interpretations are not incompatible with one-another, however, as it is perfectly possible to consider \emph{both} the average interrogator to characterize an aggregate of multiple interrogators \emph{and} all of them to be individually average~(in the sense of non-expert).
    %
}
So when Moor reframes the question underlying the Turing test as follows:

\begin{quote}
    On the average after $n$ minutes or $m$ questions is an interrogator's probability of correctly identifying which respondent is a machine significantly greater than 50 percent?~\citep[p.249]{moor1976}
\end{quote}
His reference to the ``average'' can be seen as an explicit reference to the standard methodological practice of experimental psychology since, as \citet[p.52]{tafreshi2022} tells us, ``when psychologists make inferences based on aggregate-level statistics, they make claims regarding what is true \emph{on average}''.
The underlying assumption\textemdash{}referred to as the ``nomothetic ideal''~\citep{gigerenzer1987b}\textemdash{}is that individual variability should be treated as ``error'' around a true mean which alone characterizes an objective reality.

The statistical aspect of the Turing test thus rests on an ideal of \emph{objectivity} that lies at the core of the normalist paradigm which grounds most if not all of the behavioral sciences today~\citep[see][Part I, for a comprehensive overview]{krugeretal1987b}.
Under this normalist paradigm, statistics constitute an objectivization tool which enables the elimination of subjectivity in both subjects and experimenters~\citep[see][for details]{gigerenzer1987b}.  
In the case of the Turing test, those ``experimenters'' are human interrogators which are all presumed to have partial~(and potentially partially erroneous) appreciations of what it means to be ``human'' at the individual level.\footnote{
    The unreliability of human judgment as to what should be considered human becomes even clearer in the so-called ``viva-voce'' configuration of the Turing test which involves only a single participant this time~(human or machine) being evaluated for humanness by a human interrogator~\citep[see][p.446, for the original mention]{turing1950}.
    The phenomenon of human misidentification~(humans being misidentified as machines) is what~\citet{shahandhenry2005} refer to as the ``confederate effect'' and is analyzed more extensively in~\citep{warwickandshah2015}.
}

Except that the Turing test is a test of \emph{normality} rather than humanity \emph{per se}.
That is, human interrogators are not actually required to discriminate between human and machine participants based on what they take to be ``human behavior'' but on what they take to be \emph{normal} human behavior\textemdash{}which makes all the difference.
For as~\citet[p.221]{dupre1998} tells us at the beginning of this section, the normal is also the \emph{familiar}, which explains how judgments of ``humanity'' in the Turing test can vary significantly from one judge to the next: individual human interrogators are just expressing what feels normal \emph{to them}. 
%
%
The best way to illustrate this argument is to make a quick detour by the \emph{software engineering stack exchange} and its thread dedicated to the Turing test titled ``If you could pose a question to a Turing test candidate, what would it be?''.\footnote{
    See \url{https://softwareengineering.stackexchange.com/q/64248}
}
User TZHX suggests the following answer: 

\begin{quote}
    ``That September 11th thing was amazing, wasn't it?'' \textendash{} A human would get the reference, a machine is much less likely to.~(TZHX, comment posted on April 1st 2011 at 9:27)
\end{quote}
To which user Job immediately comments:

\begin{quote}
    I believe that 70\% of Earth population~$[\ldots]$ 
    would not get that reference, and 30\% would not get it even if you say it in their native language. We are not being fair to the machines.~(Job, comment posted on April 1st 2011 at 15.39)
\end{quote}
The example clearly illustrates how we often have limited appreciations of the narrowness of our own respective subjectivities, especially when dealing with what we take to be ``normal''.
%
This justifies, in turn, the necessity to aggregate multiple interrogators' judgments so as to assess success in the Turing test.
The use of statistics in the Turing test provides a mean to move past the subjectivity of the normal as familiar\textemdash{}rooted in the singularity and potential idiosyncracies of our respective developmental experiences\textemdash{}to rely on a purportedly more objective characterization of the normal: the statistically typical.

With those considerations in mind, we can now look back at previous claims such as that of Shah and colleagues who argue that:

\begin{quote}
    One feature of `humanness' that Turing did not factor into his \emph{imitation game} for machine thinking and intelligence is that mistakes will be made by some of the human interrogators, and others are easily fooled.~\citep[p.1]{shahetal2012}
\end{quote}
For as we can understand now\textemdash{}and contrary to what is claimed\textemdash{}``mistakes'' made by human interrogators are an integral part of the Turing test. 
First, in the sense that for machines to be truly indistinguishable from human participants, human interrogators precisely \emph{need} to ``make mistakes'' and misidentify them as humans half of the time on average~(with a baseline of 50\% at least, less with a baseline of 70\%).
Second, in the sense that what motivates abstracting away from the subjectivities of individual interrogators through statistical averaging is precisely that those subjectivities are presumed to be noisy and so ``mistaken'' in the first place.\footnote{
    It is interesting to note that in later work, Warwick and Shah repeatedly misquote~\citep[p.442]{turing1950} by mentioning ``average interrogator\emph{s}'' plural rather than ``average interrogator'' singular, as in~\citep[p.2]{warwickandshah2015} or~\citep[p.1005]{warwickandshah2016a}.
    At the same time, this is not completely surprising given that their interpretation of~\citep[p.442]{turing1950} derives from their considering this notion of ``average'' to apply only to human interrogators taken individually rather than~(also) to the \emph{set} of human interrogators as a whole.
}

Having said that, it is true that Turing tells us nothing about human interrogators outside of the fact that they should not be experts about machines.
More generally, he tells us nothing about the populations of human subjects and how they should be constituted.
The question is important because it specifically conditions which~(statistical) type will be objectivized by the Turing test.
Will it be a \emph{universal} human type or, if not, at least a \emph{representative} type\textemdash{}representative of the most statistically prevalent features within the human population as a whole?
Critical psychology today warns us that if we do not pay attention to our experimental conditions and to our population samples in particular, we may end up equating what is human with what is WEIRD~\citep[that is, with what is specific to Western, Educated, Industrial, Rich and Democratic societies, to borrow the expression of][]{henrichetal2010a,henrichetal2010b}\textemdash{}especially given the prevalence of WEIRD subjects within experimental psychology in general~\citep[see also][]{arnett2008,radetal2018}.
Consequently, if we do not pay attention to our population samples in the Turing test, we may end up equating \emph{intelligent behavior} with the behavior of a \emph{particular} community of human beings\textemdash{}and most likely to equate normal human intelligence with average WEIRD intelligence. 
At its core, the argument points at the fact that human development is highly dependent on environmental contingencies\textemdash{}often both socio-culturally and socio-historially situated\textemdash{}so that there may exist a plurality of statistical types that may be objectivized within the human population as a whole~\citep{forbesetal2022}.
%
So when~\citet[p.53]{french1990} tells us that the Turing test is a test of \emph{culturally-oriented intelligence} that can only be passed ``by things that have experienced the world as we have experienced it'', we understand that this ``we'' is unlikely to refer to a representative human we, let alone a universal human we.

\subsection{Why the ``game'' aspect of the Turing test is problematic}
\label{sec:arg:game}

As we have seen so far, artificial intelligence as originally conceived by Turing is a scientific project that targets normal/average intelligence understood as a possible manifestation of normal/average human behavior. 
In this context, the Turing test undertakes to objectivize a possible human \emph{type} given how normal/average behavior corresponds to the \emph{typical} behavior of a particular community of human beings.
Artificial intelligence also fundamentally departs from artificial smartness in that it aims to objectivize ``real'' as opposed to ``ideal'' human behavior\textemdash{}how people \emph{do} behave rather than how they \emph{ought} to behave.
Turing himself distinguishes what he calls ``laws of behavior'' from ``rules of conduct'' in his original paper: 

\begin{quote}
    By `rules of conduct' I mean precepts such as `Stop if you see red lights', on which one can act, and of which one can be conscious. By `laws of behaviour' I mean laws of nature as applied to a man's body such as `if you pinch him he will squeak'~\citep[p.452]{turing1950}
\end{quote}
As such, Harnad reminds us that despite the unfortunate terminological choice\textemdash{}Imitation ``Game''\textemdash{}Turing ``means serious empirical business'':

\begin{quote}
    The game is science, the future science of cognition\textemdash{}actually a branch of reverse bioengineering.~\citep[see Harnard's comment in][p.24]{turing1950b} 
\end{quote}
Artificial intelligence as understood by Turing is therefore a scientific project that ultimately aims to objectivize the \emph{laws} or \emph{regularities} of human behavior.

Yet, the ``game'' aspect of the test still proves problematic in this respect.
%
For indeed, it puts participants in a configuration where they both have to compete to \emph{win} the game, which produces a notable side effect: it pushes human participants to behave not as they would ``normally'' do but as they think they \emph{ought} to in order to win the game and be correctly identified as humans. 
Human participants will thus strive to align with the normative ideal of normal behavior of the human interrogators as much as possible\textemdash{}or at least with what they \emph{think} this normative ideal is\textemdash{}so as not to be misidentified as machines.
If they know a lot about Shakespeare, for instance\textemdash{}to use the historical example of~\citep[p.88]{epstein1992}\textemdash{}or if they know that their behavior on a certain topic is likely to deviate from the average norm, they might be tempted to alter their behavior so as to align with that norm.\footnote{
    This point is also discussed by Sterret regarding the original configuration of the test involving woman impersonation~\citep[see][p.547, for details]{sterrett2000}.
    The ``game'' aspect of the test is indeed often approached in the literature through the question of \emph{deception} and the argument that the Turing test essentially requires machines to impersonate human beings and deceit human judges~\citep[this is precisely the argument targeted by][]{harnad1992}. 
    \citet[p.466]{pinarSayginetal2000}, for instance, explicitly tell us that ``the game is inherently about deception''. 
    %
    %
    The discussion on deception stems from the original framing of the Turing test which, before actually involving machines, requires a male participant to convincingly~(i.e. indistinguishably) impersonate a woman.
    This crucial aspect of the Turing test has been the object of much discussion~\citep[e.g.][]{genova1994b,hayesandford1995,sterrett2000,sterrett2020,traiger2000,kind2022} which I have to leave aside in the present paper for reasons of space. 
}
\begin{hide}
    \citep{sterrett2000}
    \begin{quote}
        However, the difficult task the man is set by the criterion used in the Original Imitation Game Test requires in addition that stereotypes get used for different purposes: rather than serving as common background drawn on in sincere efforts to communicate, they are to be used to mislead someone else to make inferences to false conclusions, which requires more reflection upon how others make inferences than is normally required in conversation. And, rather than relying on his well-developed cognitive habits in recognizing what an appropriate response would be, the man who takes the part of player A has to critically evaluate those recognitions; he has to go one step further and ask whether the response he knows to be appropriate for him is appropriate for a woman. \textbf{If not, he has to suppress the response he feels is appropriate for him, and replace it with the one he determines would be appropriate for a woman, or at least one that he thinks the interrogator is likely to expect of a woman. This, I think, requires a fundamentally different ability. The point that impersonation involves intellectual abilities not necessarily exhibited by the behavior impersonated is reminiscent of Gilbert Ryle's remark about a clown's impersonations: `The cleverness of the clown may be exhibited in tripping and tumbling. He trips and tumbles on purpose and after much rehearsal and at the golden moment and where children can see him and so as not to hurt himself ... The clown's trippings and tumblings are the workings of his mind, for they are his jokes; but the visibly similar trippings and tumblings of a clumsy man are not the workings of that man's mind' (Ryle, 1949, p. 33).}
    \end{quote}
\end{hide}
At the individual level, the Turing test puts human participants in a configuration where they have to compete with machines so as to appear \emph{as normal as possible} from the perspective of human interrogators.
This is precisely why Warwick and Shah recommend the following to human participants for them not to be misidentified as machines:

\begin{quote}
    Do not show that you know a lot of things\textemdash{}the judge may conclude that you are too clever to be human.
    $[\ldots]$
    Do not add new material of a different nature even if you feel this is helpful, otherwise the judge may feel that you know too much to be human.~\citep[p.11]{warwickandshah2015} 
\end{quote}
What is ultimately objectivized by the Turing test is therefore not a normal/typical human behavior corresponding to the average behavior of the population of human participants, but some \emph{expectations} of normal/average/typical human behavior of that very same population.
Which is, in turn, what can make~\citet[p.2]{goncalves2023} argue that the Turing test is ultimately a test that requires machines to imitate \emph{stereotypes} of humans. 
\begin{hide}
    \citep{goncalves2023}
    \begin{quote}
        \textbf{Essentially, according to different interpretations of the various versions of the test, the machine must be able to imitate stereotypes of a woman, a man, or a human, beside a true representative of the kind, to deceive a human interrogator about its true nature.} The new question is whether the interrogator, at a distance and having no physical contact whatsoever, would be able to distinguish the machine from the genuine individual through a conversation game. If not, the machine must be considered intelligent.
    \end{quote}
\end{hide}
The attempt to objectivize laws of behavior grounding normal human behavior thus collapses against the experimental setup of the test, as normality becomes a \emph{normative ideal} of behavior rather than a regularity under the conditions of the game.
%

\section{Conclusion}
\label{sec:conclusion}

\begin{hide}
    x. danziger? the probabilistic revolution is a revolution \emph{of} science itself rather than a revolution \emph{within} science? 

    x. meredith, statistics on steroids (maybe a bit reductory but nonetheless captures a crucial point about language models: their epistemology is that of statistics).
\end{hide}

The main goal of this paper is to show that the concept of~(statistical) normality can provide the conceptual backbone to make sense of the Turing test and weave together many of the arguments that have been carried out so far about the Turing test. 
Its core argument is that the Turing test is a test of \emph{normal intelligence} understood as~(a form of) normal/typical/average human behavior as assessed by a \emph{normal judge} which is itself an abstraction made of the aggregate of multiple interrogators' judgments. 
%
From a general perspective, this paper attempts to replace the Turing test in its historical context\textemdash{}and in its \emph{scientific} context in particular\textemdash{}to show how it can be considered a direct product of a normalist paradigm which remains very much predominant today in psychology and the behavioral sciences at large. 
Doing so, it proposes to shift our standard approach to Turing himself and to move away from the historical figure of the ``genius''~\citep[see][for explicit references]{feigenbaum1996,copelandandproudfoot1999,hilton2017} and from all the discoveries he may have anticipated or pioneered\textemdash{}such as \emph{neural networks} or \emph{connectionism}, to name a few~\citep[see][for details]{copelandandproudfoot1996,copelandandproudfoot1999}\textemdash{}to uncover all the theoretical or methodological assumptions that came before him and that he may have carried over in his practice~(willingly or not). 
The point is to look \emph{backwards} this time rather than \emph{forwards}, and to insist not so much on what singularizes Turing than on all the things that make him a regular scientist of his time. 
Far from negating or downplaying his scientific contribution, the point is simply to show that Turing can \emph{also} in many ways be considered a ``normal'' scientist after all.

\backmatter%

\bibliography{biblio}

\end{document}